\documentclass[letterpaper, 10 pt, conference]{ieeeconf}

\IEEEoverridecommandlockouts

\usepackage{cite}
\usepackage{amsmath,amssymb,amsfonts}
\usepackage{algorithmic}
\usepackage{textcomp}
\usepackage[dvipsnames, table]{xcolor}
\usepackage[T1]{fontenc}
\usepackage{scalerel}
\usepackage{siunitx}
\usepackage{comment}
\usepackage{pgffor}
\usepackage[dvipsnames]{xcolor}
\usepackage{graphicx}
\usepackage{subfigure}
\usepackage{subfloat}
\usepackage{subcaption}
\usepackage{multirow}
\usepackage[ruled,vlined]{algorithm2e}
\usepackage{bm}
\usepackage{url}
\usepackage{diagbox} 
\usepackage{balance}
\usepackage{lipsum}  
\usepackage{soul}
\usepackage{subcaption}
\usepackage{wrapfig}

\newcommand{\textcol}{black} 
\newcommand{\coloredText}[1]{\textcolor{\textcol}{#1}}

\usepackage[para]{footmisc}
\usepackage[font=small,labelfont=bf]{caption}

\makeatletter
\let\NAT@parse\undefined
\makeatother
\usepackage{hyperref}

\usepackage{eso-pic}
\newcommand\AtPageUpperMyright[1]{\AtPageUpperLeft{%
 \put(\LenToUnit{2cm},\LenToUnit{-1cm}){%
     \parbox{\textwidth}{\centering\fontsize{9}{11}\selectfont #1}}%
 }}%
\newcommand{\conf}[1]{%
\AddToShipoutPictureBG*{%
\AtPageUpperMyright{#1}
}
}

\newcommand\Tstrut{\rule{0pt}{2.0ex}}         
\def\BibTeX{{\rm B\kern-.05em{\sc i\kern-.025em b}\kern-.08em
    T\kern-.1667em\lower.7ex\hbox{E}\kern-.125emX}}

\title{\LARGE \bf Infrastructure-less UWB-based Active Relative Localization}

\author{Valerio~Brunacci, Alberto~Dionigi, Alessio~De~Angelis, and Gabriele~Costante%
\thanks{\scriptsize All the authors are with the Department of Engineering, University of Perugia, 06125 Perugia, Italy {\tt\footnotesize valerio.brunacci@studenti.unipg.it, \{alberto.dionigi, alessio.deangelis, gabriele.costante\}@unipg.it}.}%
\thanks{\scriptsize This work was supported in part by the Italian Ministry of University and Research (MUR) under Grant PRIN 2022P3JY7N, funded by the European Union - Next Generation EU.}%
}

\begin{document}
\bstctlcite{IEEEexample:BSTcontrol}
\maketitle

\conf{This work has been accepted to the IEEE/RSJ International Conference on Intelligent Robots and Systems (IROS). This is an archival version of our paper. Please cite the published version DOI: \url{https://doi.org/10.1109/IROS58592.2024.10801618}}
 


\begin{abstract}
In multi-robot systems, relative localization between platforms plays a crucial role in many tasks, such as leader following, target tracking, or cooperative maneuvering.
State of the Art (SotA) approaches either rely on infrastructure-based or on infrastructure-less setups. 
The former typically achieve high localization accuracy but require fixed external structures. The latter provide more flexibility, however, most of the works use cameras or lidars that require Line-of-Sight (LoS) to operate.
Ultra Wide Band (UWB) devices are emerging as a viable alternative to build infrastructure-less solutions that do not require LoS.
These approaches directly deploy the UWB sensors on the robots. However, they require that at least one of the platforms is static, limiting the advantages of an infrastructure-less setup.
In this work, we remove this constraint and introduce an active method for infrastructure-less relative localization. Our approach allows the robot to adapt its position to minimize the relative localization error of the other platform.
To this aim, we first design a specialized anchor placement for the active localization task.
Then, we propose a novel \textit{UWB Relative Localization Loss} that adapts the Geometric Dilution Of Precision metric to the infrastructure-less scenario.
Lastly, we leverage this loss function to train an active Deep Reinforcement Learning-based controller for UWB relative localization.
An extensive simulation campaign and real-world experiments validate our method, showing up to a 60\% reduction of the localization error compared to current SotA approaches.

\end{abstract}

\IEEEpeerreviewmaketitle



\section{Introduction} \label{sec:introduction}

Localization is a longstanding goal in robotics and is certainly one of the most crucial components of every robotic platform. 
In the past decades, it has been the object of countless research and, while \coloredText{notable} results have been achieved, it still offers important challenges.

At a high level, localization can be achieved by relying either on a fixed infrastructure or solely on onboard robot sensors. 
The former setup is implemented by coupling external sensors with onboard markers or receivers. 
This is the case of localization approaches based on Global Navigation Satellite Systems (GNSSs) in outdoor scenarios, while for indoor environments Motion Capture (MoCAP) systems \cite{merriaux2017study} can be used. 
These solutions provide highly accurate pose estimates, however, the robot is constrained within the range of the external sensor system which, in indoor scenarios, is typically limited to a single room.

Infrastructure-less solutions, on the other hand, do not require external structure at fixed known positions. 
If one is interested in the localization of a single robot, pose estimation can be achieved with respect to a known map, if available, or the global map and the robot position can be estimated jointly by exploiting the popular Simultaneous Localization and Mapping (SLAM) paradigm \cite{Bresson2017}. 
In multi-robot systems, instead, \textit{relative localization} between platforms can be exploited to refine their respective global pose estimate, or used for tasks where platforms are mainly interested in their relative pose with respect to another robot (\textit{e.g.}, leader following, target tracking or cooperative maneuvering).

\begin{figure}[t]
    \centering
    \includegraphics[width=\columnwidth]{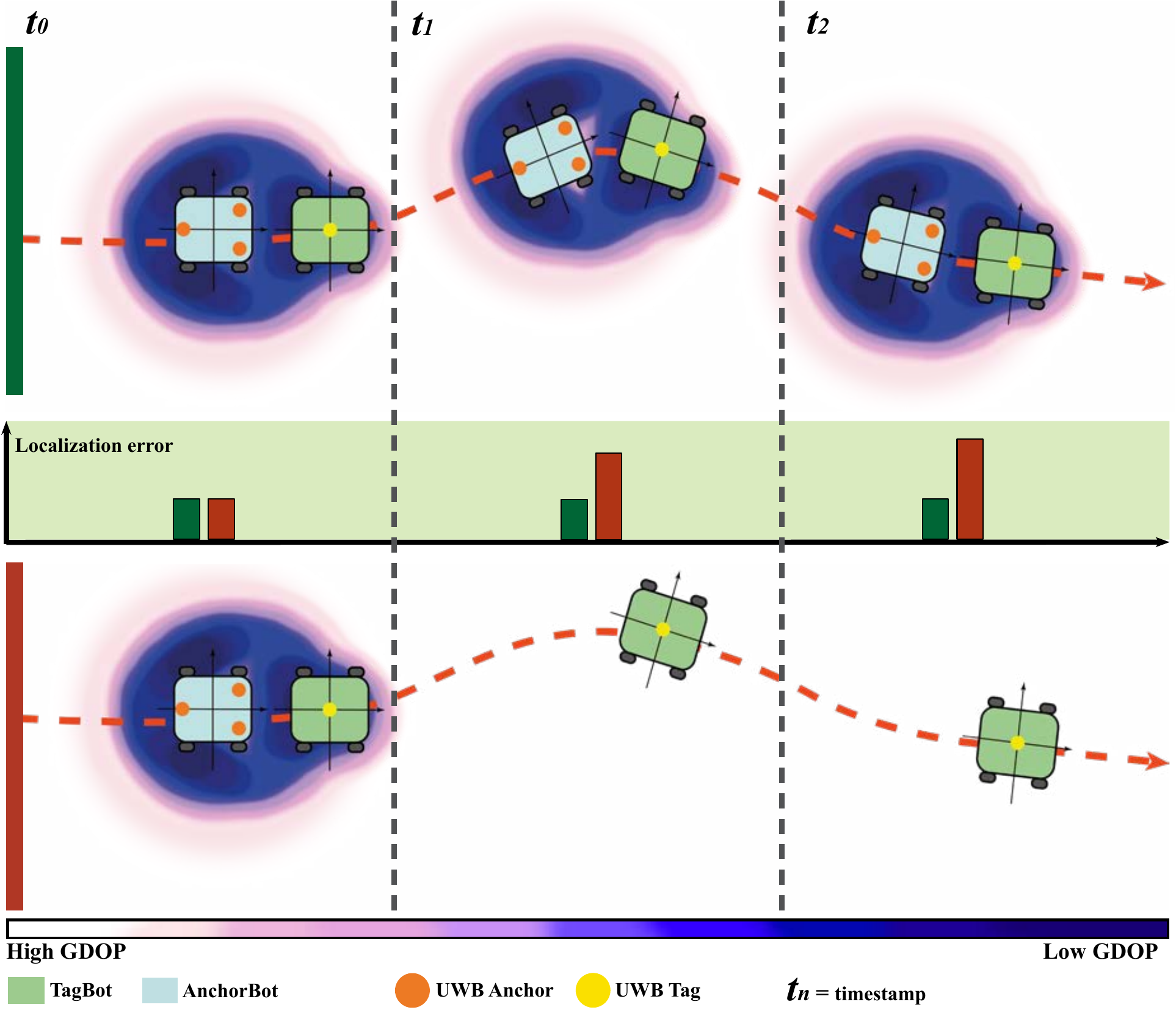}
    \caption{Overview of the proposed approach. In a UWB-based relative localization context, the performance is affected by the geometrical distribution of the sensors in the environment. Contrary to the SotA approaches (bottom figure with red bar), we develop an active method (top figure with green bar) that exploits the movement of the robot to enhance the localization performance.}
    \label{fig:overview}
   \vspace{-4ex}
\end{figure}

Relative localization can be achieved with various sensor setups, among which the most popular are based on cameras, lasers, and lidars. 
Typically, these solutions estimate the relative pose by detecting markers mounted on the robots \cite{merriaux2017study} or by inferring pose information from shape, dimensions, or texture captured by the sensors\cite{kalaitzakis2021fiducial}. 
Despite the results achieved, these approaches suffer from two important limitations: i)  through perception algorithms, which might be subject to errors and require tuning procedures; ii) they require Line-of-Sight (LoS) to operate, an assumption often violated due to occlusions and non-ideal environment conditions (\textit{e.g.}, fog or absence of illumination).

Ultra Wide Band (UWB) technology is currently flourishing as a viable alternative to achieve relative localization. 
UWB systems are cost-effective, can be used to directly access distance information between entities, and can work in Non-Line-of-Sight (NLoS) conditions. 
Most of the UWB-based solutions are infrastructure-based, \textit{i.e.}, they use fixed UWB sensors with known positions (namely the \textit{anchors}) to localize the platform equipped with another UWB device (often referred to as \textit{tag}) \cite{Zafari2019}.

Recently, UWB sensors have been also employed in infrastructure-less configurations to achieve relative localization between robots \cite{Zhou2008,Cao2021}. 
In this case, also the anchors are mounted on a mobile platform that estimates the relative position of the robot with the tag UWB. 
However, most of the State-of-the-Art (SotA) approaches keep the robot with anchors in a fixed position, limiting the advantages of an infrastructure-less setup. 
Additionally, in infrastructure-less setups the tag moves outside the convex hull delimited by the anchors. 
In practice, this translates in a worse Geometric Dilution Of Precision (GDOP) \cite{Levanon2000}, a metric that models the localization errors depending on the anchor-tag arrangement. 
In some configurations, particularly those where the tag is outside the anchor hull, small measurement errors might result in large relative position errors.

Driven by the aforementioned considerations, the purpose of this paper is to build an \coloredText{ infrastructure-less UWB-based relative 2D localization} approach for multi-robot systems that overcomes the drawbacks of SotA solutions. 
In particular, we develop an \textit{active} localization method that enables the robot with the anchors to adapt its position to reduce the relative localization error of the tag robot, see Fig \ref{fig:overview}. 
This is achieved by training a Deep Reinforcement Learning (DRL) agent that controls the platform to minimize a loss function based on the GDOP metric and on a measurement model tailored to the specific problem we consider. 
Differently from other solutions, we do not assume that the robot with the anchors is static.

The paper is organized as follows: in Section \ref{sec:related_works} we provide a literature review on the problem considered.
Then in Sections \ref{sec:preliminary} we provide the preliminary background to introduce the proposed methodology, which is described in Section \ref{sec:methodology}. 
In Section \ref{sec:experiments}, the experimental setup is outlined and the results obtained are discussed. 
Finally, in Section \ref{sec:conclusion} we draw conclusions.

\section{Related Work} \label{sec:related_works}

In the context of multi-robot systems, robot localization can be framed as either a \textit{self-localization} problem or a \textit{relative localization} one, or even as a combination of both. The former, which is also typical in single robot scenarios, aims to localize the robots with respect to a global fixed reference frame (\textit{e.g.}, a map). 
In practice, addressing this problem in multi-robot systems entails treating each platform as an independent robot.

Conversely, relative localization implies that the robots can estimate the position of each other in their respective body frames. 
This information can be used for multiple purposes: i) refining the global localization estimates of each robot; ii) localize each robot in the system with respect to a leader unit (which might perform self localization); iii) solve tasks in which global localization is not of interest, such as target tracking, or cooperative operations or maneuvers.

In the following, we focus on indoor scenarios and first provide a brief overview on infrastructure-based methodologies. 
Then, we focus on relative localization in infrastructure-less setups for multi-robot systems, emphasizing the motivations behind UWB-based strategies.
As we exploit the active localization paradigm, we also provide a literature overview of approaches based on this paradigm. 
Finally, we highlight the contribution of our work.

\textbf{Infrastructure-based Localization.}
Infrastructure-based localization (of single or multi-robot systems) in indoor environments is achieved by exploiting a fixed external structure, whose elements are static and in known positions. 
Both the structure and the mobile platforms can be equipped with either passive markers or active sensors, depending on the technology employed. 
In MoCAP systems, high-frequency cameras are used to estimate the robot's pose by tracking markers mounted on it \cite{merriaux2017study}. 
This configuration can be reverted by mounting a camera on the robot to detect visual markers installed on the walls at known positions \cite{kalaitzakis2021fiducial}.
Different solutions employ other sensors, such as radio frequency \cite{longhi2017ubiquitous,longhi2018rfid} and ultra-violet \cite{Walter2019}.

To overcome the LoS requirement of the aforementioned sensors, solutions based on UWB devices that work also in NLoS conditions have started to flourish \cite{Zhao2021}, also in combination with Inertial Measurement Units\cite{Cao2020}, cameras, and odometric sensors \cite{Zhang2023}. 

Typically, in infrastructure-based scenario, a set of UWB nodes, known as \textit{anchors}, are deployed in the corners of the room at fixed and known positions.
These devices are used to measure distances from another UWB node mounted on the mobile platform, named \textit{tag}, within the convex hull defined by the anchors. 
Thus, the localization area is confined within such a hull, and moving the structure into another environment requires considerable effort and calibration procedures.

\textbf{UWB-based Infrastructure-less Localization.}
UWB technologies can also be used in infrastructure-less scenarios. 
In this case, also the anchors are mounted on a mobile robot, and relative localization with respect to the robot housing the tag UWB sensors can be performed. While this solution is more flexible and does not require a fixed structure, it opens important challenges. 
First, the precision of the pose estimates is, in general, worse than that obtained with infrastructure-based setups. 
Secondly, the tag devices necessarily fall outside the convex hull of the anchors (which is limited \coloredText{by the robot size}). 
Thus, additional techniques need to be used to refine the pose estimate.

One of the first works that explored the infrastructure-less setup is \cite{Zhou2008}. 
However, it requires a complex and long initialization phase, reducing its practical usage. 
A viable alternative is presented in \cite{Cao2018}, although the approach proposed by the authors needs some of the UWB nodes to be static. 
In \cite{Cao2021}, both the requirements of an initialization phase and of static nodes are lifted. 
Nonetheless, they rely on additional odometry sensors and the localization performance is lower than those achieved by infrastructure-based strategies. 
Recently Gao et al. \cite{Gao2024} proposed a solution solely based on UWB sensors that is capable of providing performance comparable to those of infrastructure-based approaches. 
In this case, the anchors are deployed on the top of one robot and are arranged at the corners of an equilateral triangle with 1 meter sides. 
This arrangement provides an almost uniform GDOP distribution around the robot.
However, the platform with the anchors is kept static in the experiments, which, in practice, cancels the benefits of an infrastructure-less setup.


\textbf{Active Localization.} Considering only stationary anchors poses severe limitations on the applicability of such methods. On the other hand, in more challenging scenarios where each platform can be moved independently, an active approach could considerably enhance the localization performance in an infrastructure-less setting.
Active perception systems \cite{bajcsy1988active} take advantage of the robot motion in order to \textit{actively} acquire data from the environment and obtain more useful information with respect to the specific task of interest. 

To deal with this complex decision-making problem, recent studies are using Deep Reinforcement Learning approaches to train control policies suitable for active tasks \cite{devo2021enhancing,dionigi2022vat,chaplot2018active}, achieving remarkable results in terms of generalization, performance, and robustness. Among these, the problem of active localization was also considered. In \cite{chaplot2018active}, the authors address the global localization problem on an \textit{occupancy-grid} map with an active approach driven by DRL. The method employs RGB images for localization and the goal of the policy is to move the robot in order to acquire the views of the environment that better disambiguate its position in the map. The method shows remarkable performance on mazes, however, several assumptions limit its applicability in real-world scenarios. These drawbacks are addressed in \cite{gottipati2019deep} where the authors improve the previous approach and test their policy on a real robotic platform. Recently, the focus has shifted to active SLAM \cite{placed2023survey}, which uniﬁes the localization and the mapping problems. The goal is to actively control a robot performing SLAM to reduce both the uncertainty in its localization and the representation of the map. 

\subsection{Contributions}
While important results have been demonstrated by active approaches for localization, to the best of our knowledge there are no previous works that explored this strategy in UWB-based infrastructure-less relative localization problems. Due to the importance of the GDOP on the UWB-based systems localization performance, we propose a new framework to first study and design an anchor arrangement to adapt the GDOP to the relative localization task in mobile robot scenarios. Then we exploit this design through an active localization strategy. Specifically:

\begin{itemize}
    \item We propose an infrastructure-less active relative localization method that only leverages range measurements coming from UWB-devices. Our approach is capable of reducing the relative localization error up to 60\% with respect to current State-of-the-Art approaches;
    \item Through an extensive study on the GDOP, we (i) analyze and propose the best anchor arrangement for the \coloredText{considered} relative active localization task\coloredText{, resulting in an isosceles configuration,} and (ii) we design a novel loss function that combines GDOP and the measurement model of the UWB sensors;
    \item We use this loss function to develop a Deep Reinforcement Learning-based agent for active relative localization that controls the robot with the anchors in order to maximize the position estimation accuracy.
\end{itemize}

\section{Preliminary} \label{sec:preliminary}
\emph{Notation}:
We use lower-case symbols $x$ to denote scalars, bold lower-case symbols $\mathbf{x}$ to indicate vectors, and bold upper-case symbols $\mathbf{X}$ to denote matrices. 

\emph{Preliminary Definitions: }
In our infrastructure-less scenario, the \textit{anchors} are mounted on a mobile robot, and their positions with respect to the platform body frame $\mathcal{B}$\footnote{We omit the letter $\mathcal{B}$ in the variable definitions to simplify the notation since $\mathcal{B}$ is the only considered reference frame in this work.}, denoted as $\mathbf{p}_{a_i}=[x_{a_i},y_{a_i}]$, are known. 
On the other hand, a \textit{tag} represents the node to be localized by using distance measurements between itself and the \textit{anchors}. 
The localization is \textit{relative}, \textit{i.e.}, we are interested in estimating the tag position with respect to the body frame of the robot with the anchors. 
We use $\mathbf{p}_t=[x_t,y_t]$ to denote the true relative position of the \textit{tag}, and $\hat{\mathbf{p}}_t=[\hat{x_t},\hat{y_t}]$ to indicate its estimated relative position using noisy distance measurements.

\textbf{Geometrical Dilution of Precision -}
A comprehensive analytical characterization of GDOP is provided in \cite{Levanon2000}. 
The GDOP models the effect of geometry on the relationship between measurement error and position estimation error.
Also, it is affected by the number of anchors and by the angle between anchors and tag.
GDOP is a unitless metric, which describes how measurement errors affect the position estimation considering the geometry of the anchors.
It is defined as the ratio between the root-mean-squared error (RMSE) of the position and that of the ranging error\cite{Sahinoglu2008}:
\begin{equation}
    GDOP(\hat{\mathbf{p}}_t,\mathbf{p}_t) =\frac{\sqrt{E\{ (\hat{\beta} - \hat{\mu})^T(\hat{\beta} - \hat{\mu}) \} }}{\sigma_{range}}
    \label{eq:GDOP_Definition}
    \vspace{-2pt}
\end{equation}
where $E$ is the expectation operator, $\hat{\beta}$ is the position estimator used and $\hat{\mathbf{\mu}}$ is its mean.

In the specific case of 2D relative localization with three anchors considered in this work, an alternative definition of GDOP can be formulated as follows \cite{Levanon2000}:
\begin{equation}
\label{eq:Levanon_GDOP}
        GDOP(\mathbf{p}_t) = \sqrt{\mathbf{G}_{11}+\mathbf{G}_{22}}
\vspace{-2pt}
\end{equation}
where
\begin{equation*}
\vspace{-4pt}
        G = (\mathbf{H}^T\mathbf{H})^{-1},\qquad	
        \mathbf{H} = \begin{bmatrix}
                    \frac{x_t-x_{a_1}}{||p_{a_1} - p_t||} & \frac{y_t-y_{a_1}}{||p_{a_1} - p_t||} \\
                    \frac{x_t-x_{a_2}}{||p_{a_2} - p_t||} & \frac{y_t-y_{a_2}}{||p_{a_2} - p_t||} \\
                    \frac{x_t-x_{a_3}}{||p_{a_2} - p_t||} & \frac{y_t-y_{a_3}}{||p_{a_2} - p_t||} \\
        \end{bmatrix}
\vspace{-4pt}
\end{equation*}
\normalsize

\section{Methodology} \label{sec:methodology}
\subsection{Problem Formulation}


The objective of the proposed active relative localization method is to provide a robotic platform with a suitable control policy so that to obtain the best possible estimation of another robot position. In this paper, we consider an infrastructure-less context with two distinct robotic platforms: the first one, named \textit{AnchorBot} for brevity, is equipped with three UWB anchor nodes, and its function is to localize with respect to its body frame the second one, referred to as the \textit{TagBot}, which is equipped with a single UWB tag. The \textit{AnchorBot} have access only to the relative distances between the anchors and the tag, and the goal is to adapt its position in order to minimize the \textit{TagBot} localization error.

To this aim, we firstly conduct an in-depth anchor placement analysis \ref{sec:Sensor Displacement Analysis} to identify a configuration specifically tailored for the active localization task. Subsequently, we define a novel \textit{UWB Relative Localization Loss} \ref{sec:relative_localization_loss} that takes into account the short range measurement model of the sensors and the GDOP metric formulation. Lastly, we employ this loss to train a Deep Reinforcement Learning agent \ref{sec:Reinforcement Learning controller} that implements the control policy for the \textit{AnchorBot}.

\subsection{Anchors Displacement Analysis}\label{sec:Sensor Displacement Analysis}

\begin{figure}[t]
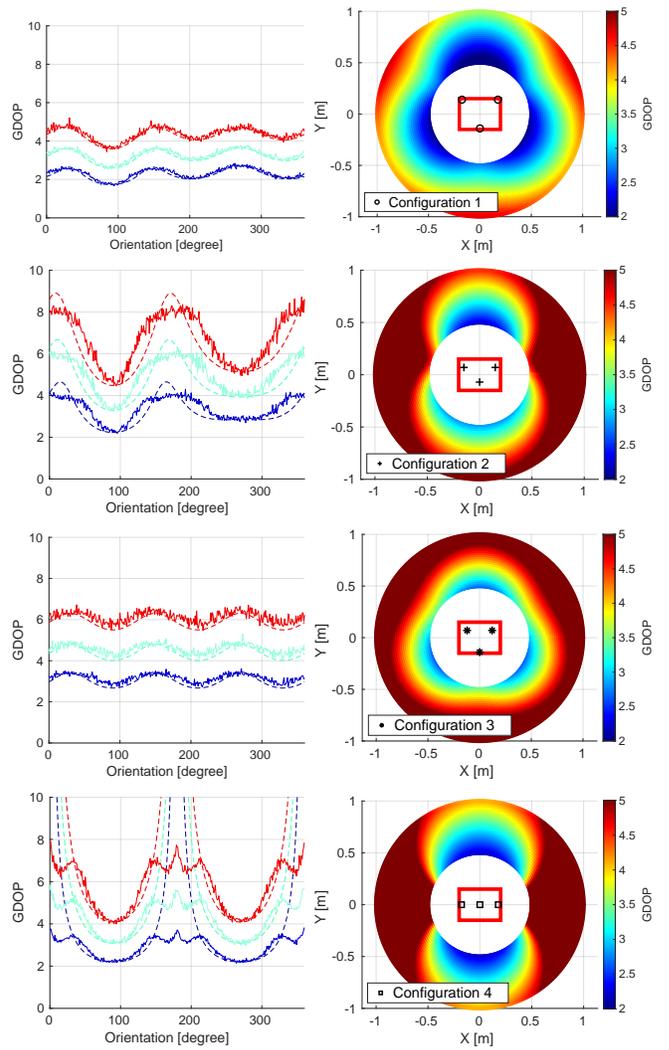

    \centering
    \foreach \i in {1,2,3,4}{
            \begin{subfigure}{}
                \includegraphics[width=0.45\columnwidth]{Methodology/_Images/GDOP/StdDev_0.05/N_mc_1000/Config_\i/Analitical_GDOP_over_angles_of_Robot_2_positions_for_Configuration_\i_with_stdDev=0.05.pdf}
                \includegraphics[width=0.52\columnwidth]{Methodology/_Images/GDOP/StdDev_0.05/N_mc_1000/Config_\i/Analitical_GDOP_values_of_the_Leader_positions_in_a_circle_of_r=0.50_for_Configuration_\i_with_stdDev=0.05.pdf}
            \end{subfigure}
        }
        \caption{Configurations from 1 to 4. On the left: simulated GDOP (continuous lines) and analytical GDOP (dashed lines) calculated on 3 circumferences of radius 50 cm (blue), 74 cm (light blue), and 1 m (red), with respect to the AnchorBot.
        On the right: analytical GDOP on a ring of points around the AnchorBot.
        }
\label{fig:GDOP_analitic}
\vspace{-4ex}
\end{figure}

As shown in Fig. \ref{fig:Loss_GDOP}, the minimum of the GDOP function falls within the convex hull delimited by the anchors. 
In an infrastructure-less scenario, however, the sensors are mounted on the AnchorBot and the anchor hull area, in practice, contains the platforms itself. To prevent collisions, the \textit{TagBot} must be positioned outside that area. 
Thus, to address this issue, new anchor configurations that can achieve a reasonably low GDOP outside the anchor hull should be explored. 
This poses the following dilemma: should we prioritize achieving a lower minimum GDOP within a specific zone, or opt for a higher but almost uniformly distributed GDOP around the AnchorBot?  \cite{Gao2024} address the latter scenario: they maintain the \textit{AnchorBot} in a stationary position, while the \textit{TagBot} moves freely. This strategy requires keeping GDOP as low as possible at all points surrounding the AnchorBot. Among various anchor configurations, the equilateral triangle arrangement (as depicted in Configuration 3 in Fig. \ref{fig:GDOP_analitic}) comes closest to achieving this goal.

However, a uniformly distributed GDOP results in a higher minimum value (\textit{i.e.}, 2.5) compared to configurations tailored for localizing the robot within specific areas (such as Configurations 1 and 2). Consequently, we select Configuration 1 (see Fig. \ref{fig:GDOP_analitic}) due to its GDOP minimum below 2. The trade-off is that the \textit{AnchorBot}’s front region hosts this minimum, requiring precise localization of the \textit{TagBot} in that zone. To address this, we adopt an active localization approach, utilizing an RL-based controller to maintain the \textit{TagBot} on the designated side of the \textit{AnchorBot}.

\noindent \textbf{Monte Carlo Simulations:}
The theoretical GDOP values described in the previous section are also validated by simulating real-world conditions using \eqref{eq:GDOP_Definition}. Specifically, we corrupt distance measurements with zero-mean Gaussian noise characterized by a variance of $5$ cm, which is representative of a realistic level of sensor noise \cite{Brunacci2023_1}. 
A Monte Carlo simulation (\coloredText{MCS}) is performed to estimate the GDOP for each configuration considered in a circular region surrounding the AnchorBot (see Fig. \ref{fig:GDOP_analitic}). 
For each sampled TagBot location in the circular region, we compute 1000 position estimates by using a trilateration technique based on the Non-Linear Least Square (NLLS) algorithm \cite{Brunacci2023_2}. Notably, our estimator is unbiased, ensuring that the mean of the estimator $\hat{\mu}$ aligns with the true parameter \cite{Sahinoglu2008}.

\subsection{UWB Relative Localization Loss} \label{sec:relative_localization_loss}

As anticipated in the previous Section, we choose an anchor configuration that attains a minimum GDOP in front of the AnchorBot. We assign the task of controlling the platform to maintain the TagBot within this minimum area (thus minimizing the relative localization error) to an active control approach. To this aim, we chose to adopt a learning-based approach for the controller design and, in order to set up the optimization process, we design a loss function encoding the requirement of the task considered.

In an UWB-based localization context, directly using the GDOP as the loss function might seem a reasonable choice since, by definition, it is a performance metric for positioning. 
However, this is impractical since, in an infrastructure-less scenario, the GDOP minimum falls in the AnchorBot footprint area, as described in \ref{sec:Sensor Displacement Analysis}.
Furthermore, another aspect that influences localization performance is distance. More specifically, as observed in \cite{Iacono2022,Polonelli2022}, UWB sensors are not able to provide a reliable measurement for distances less than 10 cm, and for short-range measurements, typically between 10 and 25 cm, they exhibit greater errors in terms of measurement accuracy and variance compared to what they are able to provide for distances from 25 cm up to 5 m. 

Motivated by the above considerations, we develop a novel \textit{UWB Relative Localization Loss} $\textstyle \ell(\mathbf{p}_t): \mathbb{R}^2 \rightarrow \mathbb{R}$ that combines the GDOP metric with a suitable measurement model for the anchors and takes into consideration the UWB performance degradation for short ranges. The loss function depends on the tag position $\mathbf{p}_t$ and is defined as:
\begin{equation}\label{eq:loss}
\vspace{-4pt}
    \ell(\mathbf{p}_t)= GDOP(\mathbf{p}_t) + \alpha \sum_{i=0}^{N}f_e(||\mathbf{p}_t-\mathbf{p}_{a_i}||)
\vspace{-2pt}
\end{equation}
where $\mathbf{p}_{a_i}$ is fixed and constant in the formula, $N$ is the number of \textit{anchors}, and $\alpha$ is a tunable hyper-parameter. The $GDOP(\mathbf{p}_t)$ term corresponds to the formulation in \eqref{eq:Levanon_GDOP}, and $f_e$ is the analytical function of the measurement model derived from the study in \cite{Iacono2022}:
\begin{equation}\label{eq:measurement_model}
    f_e(||\mathbf{p}_t-\mathbf{p}_{a_i}||)=c_1e^{k_1 ||\mathbf{p}_t-\mathbf{p}_{a_i}||}
\end{equation}
where $c_1$ and $k_1$ are two parameters with value 0.51 and -3.152 1/m, respectively.

\begin{figure}[t]
    \centering
            \subfigure[]{
                \includegraphics[width=0.46\columnwidth]{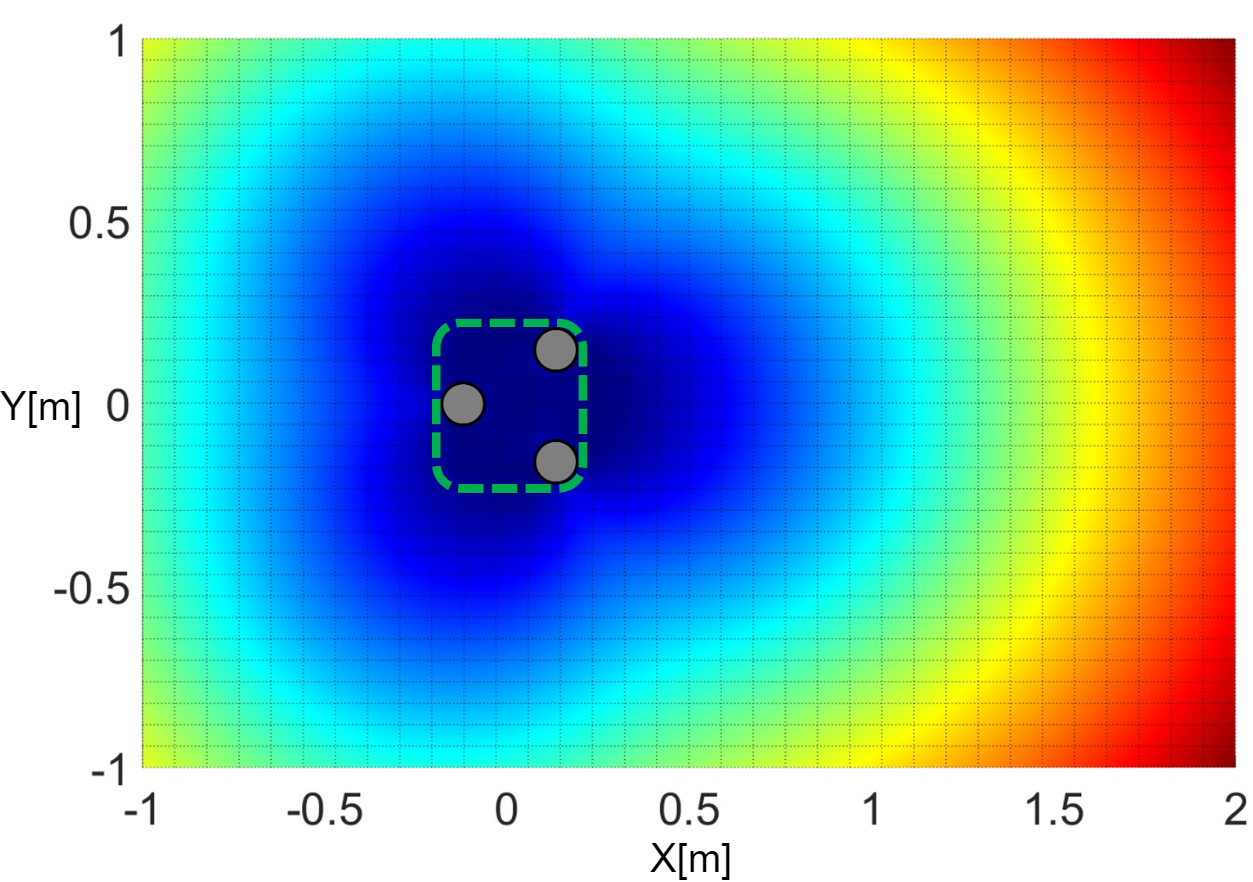}
                \label{fig:Loss_GDOP}
            }
            \subfigure[]{
                \includegraphics[width=0.46\columnwidth]{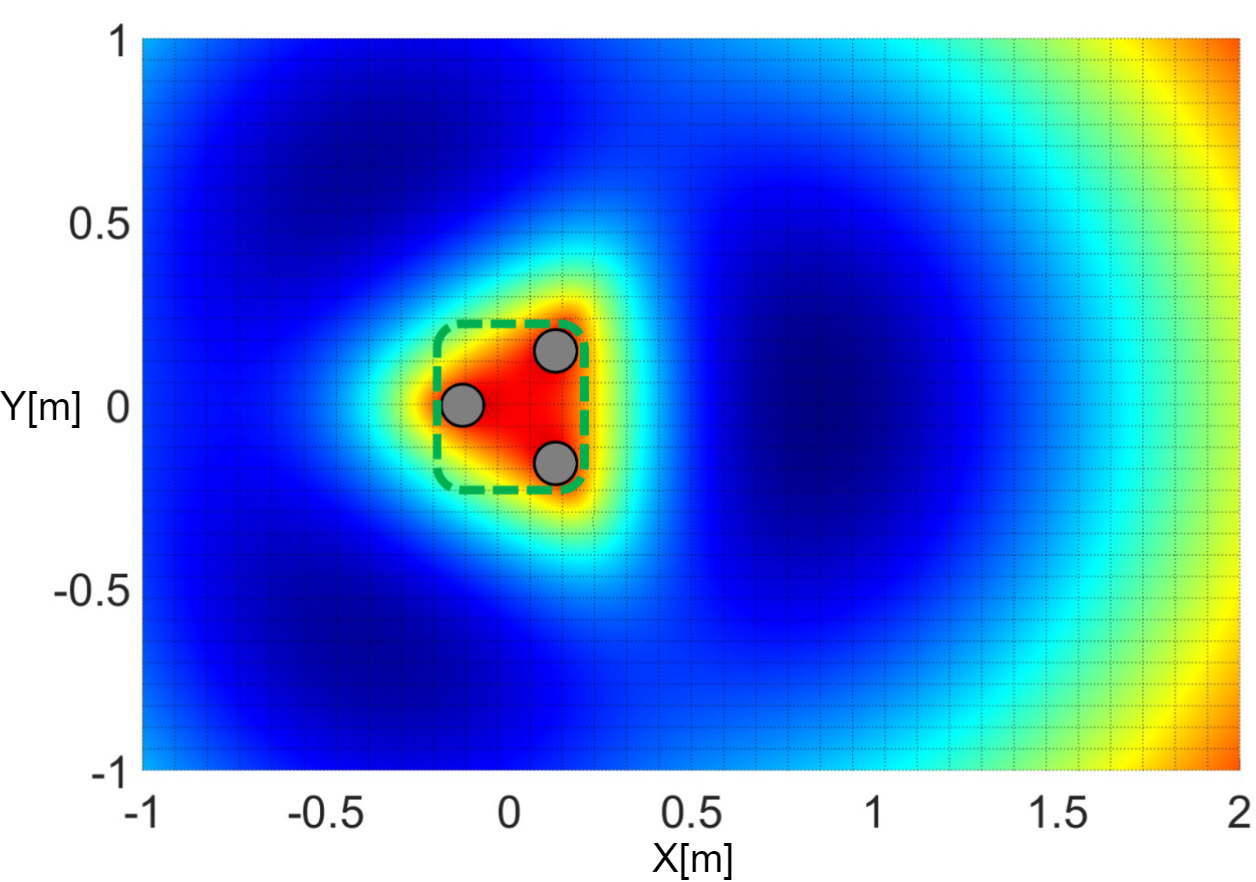}
                \label{fig:Loss_UWB}      
            }
            \caption{Comparison between (a) the GDOP and (b) the UWB Relative Localization Loss. The grey dots are the anchors, while the dashed green line represents the footprint of the \textit{AnchorBot}. Cold colors indicate low values, and warm ones refer to high values.}
\vspace{-4ex}
\end{figure}

In Fig. \ref{fig:Loss_UWB} we show that the developed loss function $\ell(\mathbf{p}_t)$ has a suitable global minimum in front of the robot, which is unique and takes into account the performance drop in the proximity of the anchors.  


\subsection{Deep Reinforcement Learning controller}\label{sec:Reinforcement Learning controller}

In order to develop a control policy to effectively address the Active Relative Localization task, we employ a learning-based strategy and define a Reinforcement Learning problem.

\textbf{Simulation Environment - } Learning a control policy directly on a real platform is not, in general, feasible due to safety and temporal constraints. Therefore, it is necessary to employ a simulation environment to train the algorithm.
In this work, we model the \textit{AnchorBot} as a unicycle-type differential drive platform \cite{leomanni}.
The robot can move along its heading direction with linear velocity $v$ and spin around the center point position of the vehicle with angular velocity $\omega$.
The commands $v$ and $\omega$ are provided by the controller and correspond to the input of the robot. Moreover, the sampling instants are denoted by $k=j t_s$, in which $j$ is the discrete-time step number and $t_s$ is the sampling time.

On the other hand, during training, the \textit{TagBot} moves along a sinusoidal trajectory parameterized with respect to \textit{TagBot} starting position as follows:
\begin{equation}\label{referencetrajectory}
\textbf{p}_t(k) = 
\begin{bmatrix}
    A_x  \sin(2\pi f_x  k + \phi_x)\\ 
    A_y  \sin(2\pi f_y  k + \phi_y)\\ 
\end{bmatrix}
\end{equation}
where ($A_x$, $A_y$), ($f_x$, $f_y$) and ($\phi_x$, $\phi_y$) are respectively the amplitude, the frequency and the phase along the $(x,y)$ axes.

\textbf{DRL Strategy - }In our settings, the two robots are free to move on a 2-D plane. The \textit{AnchorBot} controlled by the DRL agent is equipped with three UWB \textit{anchors} that respectively return their distances $\mathbf{d}(k) = \left[d_1(k), d_2(k), d_3(k)\right]$ from the \textit{tag} mounted on the \textit{TagBot}. The observation $\mathbf{o}(k)$ is composed of $H$ consecutive readings of the sensors
\begin{equation}
\label{eq:observation}
    \mathbf{o}(k) = [\mathbf{d}(k), \mathbf{d}(k - 1),  \cdots,  \mathbf{d}(k - H + 1)]
\end{equation}
resulting in a vector of length $3 \times H$. On the other hand, since the robot can be moved by specifying the desired linear velocity $v$ and angular velocity $\omega$, the action space is a continuous control command in the form of $\mathbf{u}(k) = \left[v(k),\omega(k)\right]$.

Given the large dimensionality of the state space and the continuous nature of the considered problem, employing a classical tabular RL approach is not feasible.
Instead, we employ a DRL strategy, leveraging Deep Neural Network (DNN) approximators. In particular, we adopt an \textit{asymmetric actor-critic} framework \cite{dionigi2022vat} and we design two DNN architectures: the \textit{actor} (A-DNN) one learns the optimal policy $\pi(o(k))$, while the \textit{critic} (C-DNN) counterpart is used during training phase only to evaluate such a policy.

The A-DNN consists of a Multi-Layer Perceptron (MLP) with three hidden layers, each one characterized by $256$ neurons and ReLU activations. The input to the network is the flattened vector $\mathbf{o}(k)$ in ~\eqref{eq:observation}, and its output $\mathbf{u}(k)$ contains the linear and the angular velocities needed to control the robot. Furthermore, to impose the physical constraints that come from the saturation limits of the motors, we add a $\tanh$ function to the last layer of the network in order to contain the output of the network within a reasonable range.

Through the use of an asymmetric framework, we have the flexibility to provide the C-DNN with more privileged information compared to what is accessible to the A-DNN during inference. Consequently, in order to develop a more robust criterion for policy evaluation, we define the critic observation $o_{c}(k)=\!\left[\mathbf{p}_t(k), \theta(k) \right]$\coloredText{, where $\theta(k)$ is the \textit{AnchorBot} orientation}.
The architecture of the C-DNN is an MLP composed of 3 hidden layers with $256$ neurons each and ReLU activations. The network takes as input the concatenation of $\mathbf{o}_c(k)$ with $\mathbf{u}(k)$, and outputs estimation of the \textit{action-value} function $Q_\mathbf{\pi}(\mathbf{o}_c(k), \mathbf{u}(k))$.

\textbf{Optimization Procedure - }
Given the continuous formulation of the problem, we decided to train the A-DNN and the C-DNN by employing the Soft Actor-Critic (SAC) \cite{haarnoja2018soft} RL-framework, which is one of the most recent off-policy strategies. The goal of the reward signal $r(k)$ is to minimize the localization error while avoiding collisions and reducing the control effort. To this purpose, we firstly normalize the UWB-Localization Loss so that its minimum value is zero:
\begin{equation} \label{eq:scaled_loss}
    \ell_{scaled}(\mathbf{p}_t(k)) = \frac{\ell(\mathbf{p}_t(k)) - \ell_{min}}{\ell_{max} - \ell_{min}},
\end{equation}
where $l_{min}$ and $l_{max}$ are the minimum and maximum values of the loss function, respectively. Subsequently, we design a first reward term that optimize \eqref{eq:scaled_loss}: 
\begin{equation} \label{rew:loss}
    r_{l}(k) = \max\left(0, 1 -\ell_{scaled}(\mathbf{p}_t(k))\right),
\end{equation}
It can be noted that $r_{l}(k)$ is 1 only when the \textit{AnchorBot} manages to keep the \textit{TagBot} in the point that minimizes \eqref{eq:loss}.

In order to also \coloredText{consider} the control effort on the actuators, a reward \coloredText{penalty} contribution $r_{u}$ is included in the design:
\begin{equation} \label{pen:effort}
    r_{u}(k) = \dfrac{\| u(k) \|}{1 + \| u(k) \|}.
\end{equation}

The reward function is obtained by adding up the above terms. Furthermore, to ease the learning process, we removed the local minima of \eqref{rew:loss} by considering its contribution only when the \textit{TagBot} is in front of the \textit{AnchorBot} within a suitable angle, which results in:
\begin{equation} \label{rew:total}
r(k) = \left\{\begin{array}{l c}
\!\!r_{l}(k) \!- \! k_u r_{u}(k) & x_t(k)>0 \And |\gamma|<\frac{\pi}{4} \\
\!\!-k_c & \| p_t(k) \| < d_{m} \\
\!\!-k_u r_{u}(k) & \textit{otherwise}
\end{array}\right.
\vspace{-4pt}
\end{equation}
with
\small
\begin{equation}
    \gamma = \arctan \frac{y_t(k)}{x_t(k)}
\end{equation}
\normalsize
where $k_u>0$ is a weighting parameter that balance the performance and the control effort terms, and $k_c$ is a large positive constant that penalize the RL agent when it does not meet the collision avoidance constraint in which $d_{m}$ is the minimum distance allowed between the two robots.

\section{Experiments}\label{sec:experiments}
\begin{table*}[!t]
\tiny
\centering
\caption{Results of the simulation experiments. The mean $\mu$ and the standard deviation $\sigma$ of the \coloredText{MLE} in cm are reported.}
\label{tab:results}
\resizebox{2\columnwidth}{!}{%
\begin{tabular}{c|c|c|c|c|c|c|c|c|c|c|c|c|c}
\cline{3-14}
\centering \multirow{3}{*}{\shortstack{\textit{AnchorBot} - \textit{TagBot}\\ Init. Distance}} & \multirow{3}{*}{Method} & \multicolumn{6}{c|}{Static \textit{TagBot}} & \multicolumn{6}{c}{Dynamic \textit{TagBot}}\Tstrut\\
\cline{3-14}
&& \multicolumn{2}{c|}{Front Spawn} & \multicolumn{2}{c|}{Side Spawn} & \multicolumn{2}{c|}{Behind Spawn} & \multicolumn{2}{c|}{Straight Line} & \multicolumn{2}{c|}{Circle} & \multicolumn{2}{c}{Square} \Tstrut\\
\cline{3-14}
&& $\mu$ & $\sigma$ &$\mu$ & $\sigma$ & $\mu$ & $\sigma$ &$\mu$ & $\sigma$ & $\mu$ & $\sigma$ &$\mu$ & $\sigma$ \\
\hline
\centering \multirow{4}{*}{70 cm} & SUL-EQ \cite{Gao2024}
& 11.0 & 7.8 
& 12.9 & 9.6 
& 16.0 & 13.9  
& 28.2 & 24.9  
& 13.3 & 10.4  
& 21.6 & 18.2 \Tstrut\\ 
& SUL-IS
& 10.3 & 7.1 
& 12.9 & 9.7 
& 21.6 & 30.7  
& 26.0 & 22.8  
& 13.5 & 14.1  
& 20.2 & 16.9 \\ 
& AUL-EQ
& 11.5 & 8.1  
& 11.7 & 8.3  
& 11.5 & \textbf{9.3}  
& 12.7 & 9.1  
& 11.4 & 8.0  
& 11.9 & 8.5 \\ 
& AUL-IS (Our)
& \textbf{10.7} & \textbf{7.5}  
& \textbf{11.0} & \textbf{7.8}  
& \textbf{11.3} & 9.7  
& \textbf{11.7} & \textbf{8.3}  
& \textbf{10.6} & \textbf{7.4}  
& \textbf{11.1} & \textbf{7.8} \\ 
\hline
\multirow{4}{*}{85 cm} & SUL-EQ \cite{Gao2024}
& 13.4 & 9.7 
& 15.4 & 11.6 
& 18.1 & 14.1  
& 30.8 & 26.6  
& 15.7 & 11.8  
& 23.8 & 19.9 \Tstrut\\ 
& SUL-IS
& 12.5 & 8.9  
& 15.4 & 11.5  
& 19.4 & 22.9  
& 28.4 & 24.4  
& 15.5 & 13.4  
& 22.2 & 18.4 \\ 
& AUL-EQ 
& 11.6 & 8.3  
& 12.1 & 8.8  
& 12.2 & \textbf{9.1}  
& 12.7 & 9.1  
& 11.9 & 8.5  
& 12.0 & 8.7 \\ 
& AUL-IS (Our)
& \textbf{10.9} & \textbf{7.7}  
& \textbf{11.4} & \textbf{8.3}  
& \textbf{11.7} & 9.3  
& \textbf{11.8} & \textbf{8.4}  
& \textbf{11.1} & \textbf{7.8}  
& \textbf{11.8} & \textbf{8.4} \\ 
\hline
\multirow{4}{*}{100 cm} & SUL-EQ \cite{Gao2024}
& 15.9 & 11.7 
& 18.2 & 13.7 
& 20.7 & 15.7  
& 33.3 & 28.3  
& 18.3 & 13.7  
& 26.2 & 21.6 \Tstrut\\ 
& SUL-IS
& 14.9 & 10.7  
& 18.0 & 13.6  
& 20.2 & 19.5  
& 31.0 & 26.2  
& 17.7 & 14.1  
& 24.4 & 20.0 \\ 
& AUL-EQ 
& 12.0 & 8.7  
& 12.6 & 9.5  
& 12.9 & \textbf{9.8}  
& 13.1 & 9.5  
& 12.2 & 8.7  
& 12.3 & 9.0 \\ 
& AUL-IS (Our)
& \textbf{11.2} & \textbf{8.0}  
& \textbf{11.9} & \textbf{9.0}  
& \textbf{12.3} & \textbf{9.8}  
& \textbf{12.1} & \textbf{8.7}  
& \textbf{11.3} & \textbf{7.9}  
& \textbf{12.1} & \textbf{8.7} \\ 
\hline
\multirow{4}{*}{150 cm} & SUL-EQ \cite{Gao2024}
& 24.6 & 18.6 
& 26.8 & 20.3 
& 29.3 & 22.3  
& 42.1 & 34.8  
& 26.9 & 20.4  
& 34.3 & 27.6 \Tstrut\\ 
& SUL-IS
& 22.8 & 17.1  
& 26.9 & 20.4  
& 27.5 & 22.0  
& 39.0 & 31.8  
& 25.9 & 19.8  
& 31.9 & 25.5 \\  
& AUL-EQ 
& 15.0 & 12.2  
& 16.2 & 13.7  
& 17.0 & 14.4  
& 17.2 & 13.7  
& 12.8 & 9.4  
& \textbf{14.5} & \textbf{12.1} \\  
& AUL-IS (Our)
& \textbf{13.9} & \textbf{11.2}  
& \textbf{15.4} & \textbf{13.1}  
& \textbf{16.0} & \textbf{13.7}  
& \textbf{16.0} & \textbf{12.6}  
& \textbf{11.8} & \textbf{8.5}  
& 16.0 & 12.6 \\  
\hline
\multirow{4}{*}{200 cm} & SUL-EQ \cite{Gao2024}
& 33.2 & 25.2 
& 35.6 & 27.1 
& 37.9 & 29.0  
& 50.7 & 41.0  
& 35.6 & 27.1  
& 42.8 & 33.9 \Tstrut\\ 
& SUL-IS
& 30.8 & 23.1  
& 35.6 & 27.1  
& 35.3 & 27.4  
& 47.0 & 37.7  
& 34.3 & 26.2  
& 39.7 & 31.3 \\  
& AUL-EQ 
& 21.0 & 17.9  
& \textbf{22.7} & \textbf{19.5}  
& 24.5 & 20.6  
& 25.4 & 20.1  
& 13.3 & 10.3  
& 17.5 & 16.6 \\  
& AUL-IS (Our)
& \textbf{19.5} & \textbf{16.6}  
& 23.9 & 20.4  
& \textbf{22.5} & \textbf{19.3}  
& \textbf{23.8} & \textbf{18.6}  
& \textbf{12.2} & \textbf{9.4}  
& \textbf{15.9} & \textbf{15.1} \\  
\hline
\end{tabular}%
}
\vspace{-1mm}
\vspace{-4pt}
\end{table*}
\subsection{Implementation details}\label{sec:implementation_details}
Based on the study carried out in \ref{sec:Sensor Displacement Analysis}, we opt for a configuration of the anchors on the \textit{AnchorBot} over the vertices of an isosceles triangle, whose positions with respect to the robot body frame are $(0.14 m,0.175 m)$, $(0.14 m,-0.175 m)$ and $(-0.14 m,0 m)$, respectively. Hence, in the following discussion our approach is referred to as \textbf{AUL-IS} (\textbf{A}ctive \textbf{U}WB-\textbf{L}ocalization - \textbf{IS}oscele triangle configuration).
On the other hand, the tag to be localized is considered to be in the center point of the \textit{TagBot}.

The hyper-parameter $\alpha$ of the UWB Relative Localization Loss \eqref{eq:loss} is a design parameter that balances the GDOP metric and the short-range measurement model terms.
In order to obtain a position for the minimum that guarantees high localization performance, we set its value to 10.

For the optimization of the active controller, we use the Stable-Baselines3 \cite{stable-baselines3} implementation of SAC, which we properly customized to implement our \textit{asymmetric actor-critic} structure. 
The A-DNN and C-DNN have been trained by employing the Adam optimizer with a learning rate of 0.0003, a discount factor $\gamma$ of 0.99, and a batch size of 256.
The training phase is structured in episodes with a maximum duration of 30 s. 
At the beginning of each one, the \textit{TagBot} is randomly spawned with a uniform distribution at a distance between 60 cm and 100 cm from the \textit{AnchorBot}. 
In addition, to achieve robustness against the UWB measurement noise, we corrupt the distance data with an additive zero-mean Gaussian noise with a variance 5 cm, according to \cite{Brunacci2023_2}. 
During the episode, the \textit{TagBot} remains stationary for the first 10 seconds and then starts moving along the trajectory reported in \eqref{referencetrajectory}.
Furthermore, to produce a different \textit{TagBot} movement for each episode, we uniformly randomize the amplitude, the frequency and the phase of the trajectory in the intervals $\left[1, 2.5\right]$m, $\left[0.008, 0.016\right]$Hz, and $\left[0, 2\pi \right]$rad, respectively. 
The episode terminates when the maximum number of steps is reached or if the \textit{AnchorBot} violates the collision avoidance constraint. 

The training process requires about 10 minutes and 1.1GB of VRAM on a workstation equipped with NVIDIA Quadro GV100. 
At inference time, the VRAM required for the Actor Network is about 320 kB, and the time required to compute the action is approximately 1 ms.
\subsection{Experimental Setup} \label{sec:exp_setup}
The proposed \textbf{AUL-IS} approach is tested through an extensive simulation campaign and validated with real-world experiments on a wheeled robotic platform. More specifically, we perform two distinct types of experiments. In the first one, the \textit{TagBot} remains stationary and assumes three different positions: in front of, to the side, and behind the \textit{AnchorBot}. The second set of tests is, instead, aimed at assessing the capability of the proposed active method to localize a moving platform. Specifically, we evaluate the generalization capabilities of the RL controller by considering trajectories with shapes different from those used for training, \textit{i.e.}, a two-meters straight line, a circular trajectory centered in the \textit{AchorBot}, and a one-meter side rectangular shape.
To ensure a comprehensive experimental campaign, we vary the spawn distances for the \textit{TagBot} relative to the \textit{AnchorBot}, ranging from 70 cm up to 200 cm. For each experiment, we run a total of 1000 realizations.

As a comparison baseline, we employ the method proposed in \cite{Gao2024}. Following the original work, the \textit{AnchorBot} is kept stationary and the anchors are mounted on the vertices of an equilateral triangle. We refer to this method as \textbf{SUL-EQ} (Static UWB Localization - Equilateral configuration). Moreover, to highlight the advantages of our design choices, we also included as baselines (i) a stationary \textit{AnchorBot} with our isosceles triangle configuration for the anchors (\textbf{SUL-IS}) and (ii) an active version of \cite{Gao2024} in which the anchors have an equilateral triangle configuration, but the \textit{AnchorBot} is controlled by our DRL active policy (\textbf{AUL-EQ}).

To evaluate the performance of the proposed approach and the baselines, we compute the  \coloredText{Mean Localization Error (MLE)} 
\coloredText{$\textstyle  \frac{1}{N}\sum_{t=1}^{N} || \hat{\mathbf{p}_t}-\mathbf{p}_t||$}
over multiple runs of the experiments, and we adopt as comparison metrics the mean $\mu$ and standard deviation $\sigma$.
Moreover, since the only available information are the three distance measurements between the \textit{anchors} and the \textit{tag}, we employ the  NLLS  method to obtain the position estimation $\mathbf{\hat{p}}_t$ of the \textit{TagBot}, as for the \coloredText{MCS} in \ref{sec:Sensor Displacement Analysis}. It is important to highlight that the proposed method is independent from the localization algorithm, since the DRL controller directly employs the raw distances coming from the UWB devices to move the robot.

    
\subsection{Experimental Results} \label{sec:exp_results}

\begin{figure}[t]
    \centering
    \includegraphics[width=\columnwidth]{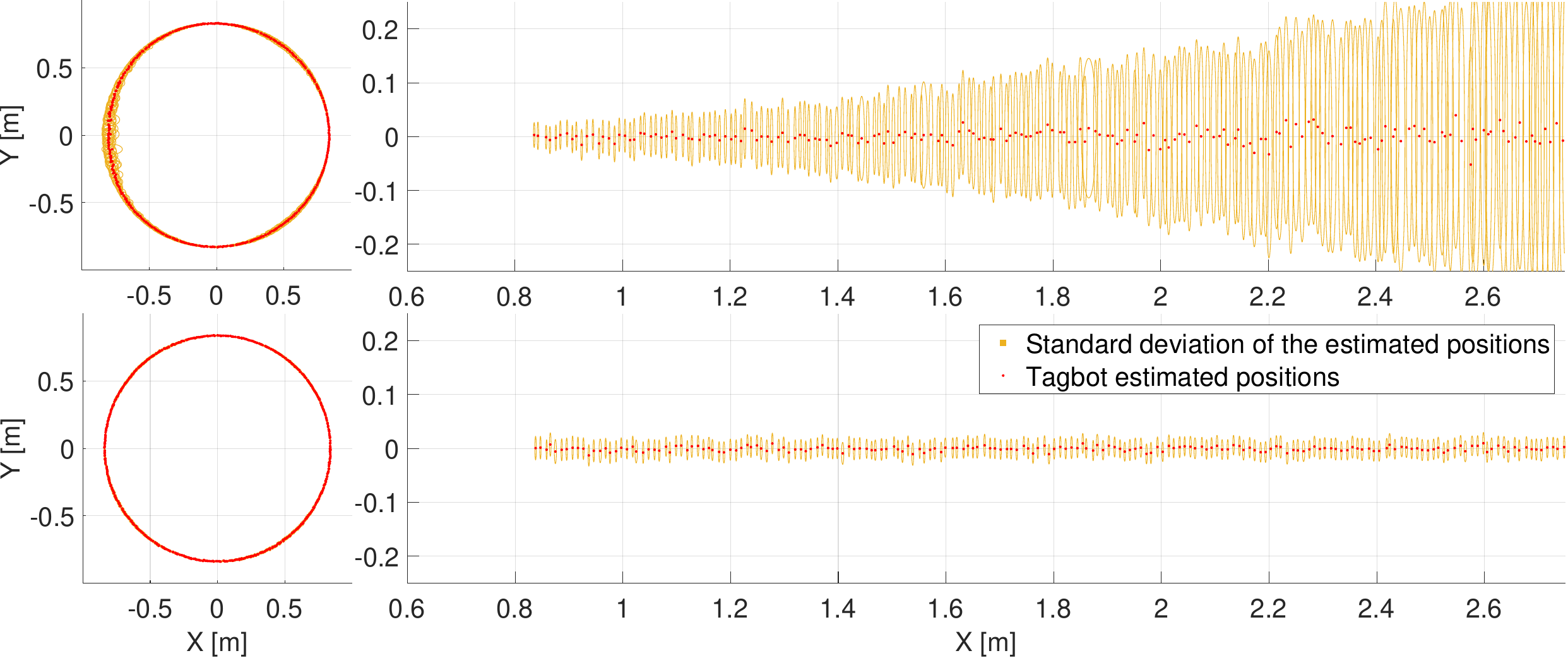}
\caption{\coloredText{Plots of the dynamic experiments for both \textbf{SUL-EQ} (top) and \textbf{AUL-IS} (bottom) are shown for the \textit{Straight Line} and \textit{Circle} experiments.
}}
\label{fig:plots_Simulations}
\vspace{-4ex}
\end{figure}

The results of the simulated experimental campaign are reported in Table \ref{tab:results}. 
A first key finding is that the active approaches \textbf{AUL-EQ} and \textbf{AUL-IS} outperform their static counterparts \textbf{SUL-EQ} and \textbf{SUL-IS} in every situation, achieving an error reduction up to 60\% (Straight Line-150 cm). 
This result is a direct consequence of the control policy developed by the DRL agent, which is able to effectively move and orient the \textit{AnchorBot} so as to maintain the \textit{TagBot} in low GDOP areas. 
Moreover, it can be observed that the specialized anchor configuration exploited by \textbf{AUL-IS} can further enhance the localization performance. 
Clearly, as demonstrated by the static \textit{TagBot} experiments and in accordance with the findings in \cite{Gao2024}, an equilateral triangle disposition is the best choice with a stationary \textit{AnchorBot}. 
However, when we leverage an active approach that can take advantage of the \textit{AnchorBot} movements, it is convenient to use an isosceles triangle configuration that places the minimum of the GDOP to be in front of the \textit{AnchorBot}.

A second important aspect to emphasize is that \textbf{AUL-IS} is able to make localization performance independent from the \textit{TagBot} position and movements, and from the \textit{AnchorBot} initialization. 
This capability is not observed in \textbf{SUL-EQ} or \textbf{SUL-IS} that do not include strategies of adaptation to GDOP conditions. 
For instance, if we refer to the second row of Table \ref{tab:results} (\textit{i.e.,} \textit{AnchorBot} - \textit{TagBot} Init. distance 85 cm) and we consider both static and dynamic experiments, the difference between the worst ($\mu=30.8$) test and the best ($\mu=13.4$) one achieved by \textbf{SUL-EQ}  is about 17.4 cm, while for \textbf{AUL-IS} this gap is less than 1 cm. A further demonstration is shown in Fig. \ref{fig:plots_Simulations}. 
It can be observed that for \textbf{SUL-EQ}, the localization error variance increases with respect to the orientation and the distance between the two robots. 
On the other hand, \textbf{AUL-IS} achieves remarkable performance in all the considered scenarios and the variance of its localization error is always low and constant.


\begin{figure*}[t]
	\centering
	\includegraphics[width=\textwidth]{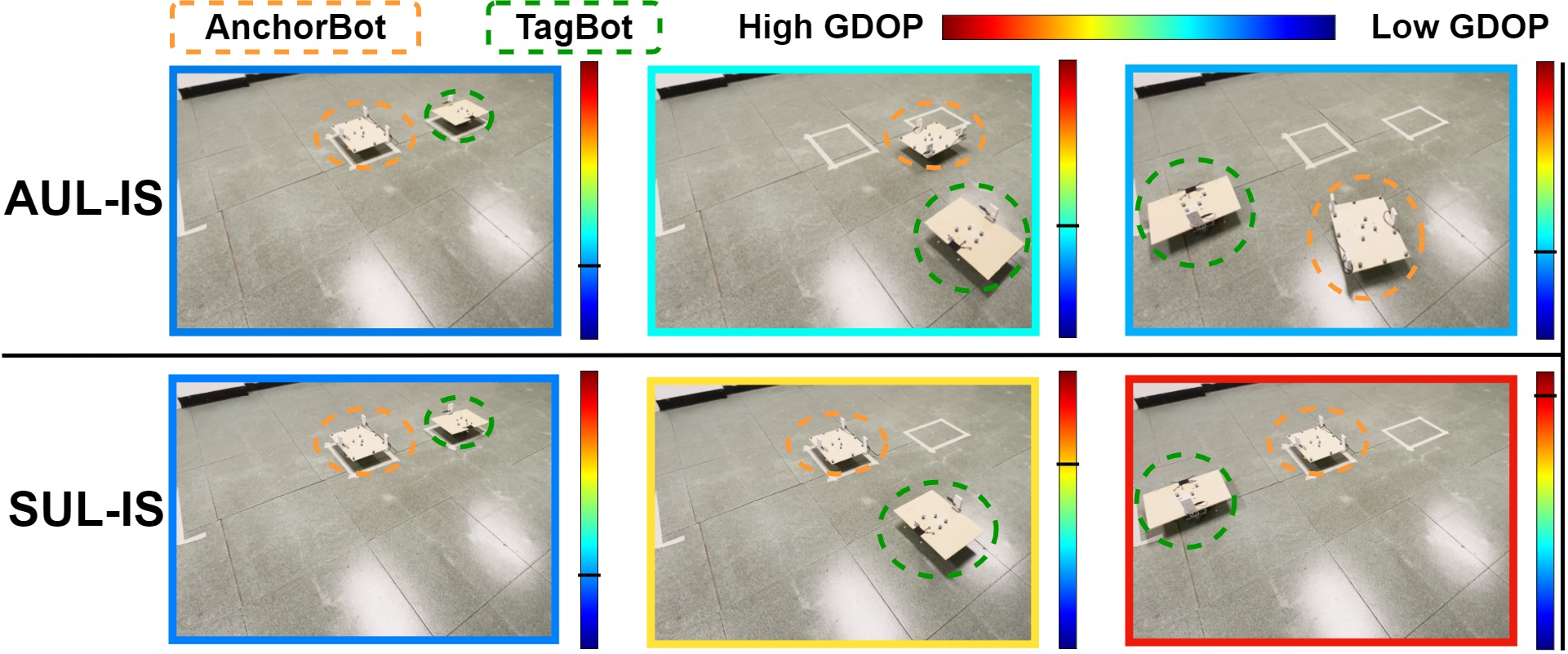}
	\caption{Circle dynamic experiment with a radius of 85 cm in real-world. The colored outlines in the figures indicate the GDOP value for the considered timestamp according to the colorbar scale.  }
	\label{fig:realEnvs}
\end{figure*}

\subsection{Real World Validation}\label{real_world_validation}
Real-world experiments are carried out to validate the results obtained in the simulation campaign and to demonstrate the transferability of the DRL controller on a real platform. 
To this aim, we employ two GoPiGo3\footnote{\url{https://www.dexterindustries.com/gopigo3/}} robots and four Quorvo DWM1001-Dev UWB devices \footnote{\url{https://www.qorvo.com/products/d/da007946}} configured as described in \ref{sec:implementation_details}. 
The three \textit{anchors} and the \textit{tag} are placed on a wooden board screwed to the frame of the \textit{AnchorBot} and the \textit{TagBot}, respectively. Both robots are equipped with an embedded platform for the computation of the control algorithm and distance measurement acquisition.

The experiments are performed indoors, and the OptiTrack\footnote{\url{https://optitrack.com/}} motion capture system is employed to obtain ground truth information. 
We test our method with either stationary and dynamic \textit{TagBot} conditions, and in every configuration described in \ref{sec:exp_setup}. 
Moreover, without loss of generality, we considered 85 cm as the \textit{TagBot} start distance, which was the largest possible to perform experiments within the area covered by the MoCAP system. 
We chose \textbf{SUL-IS} as the comparison baseline since in simulation it shows the best overall performance with respect to \textbf{SUL-EQ} in the 85 cm scenarios. 
The experiments are averaged over three different runs of about 30 seconds each, and before performing the actual trial, we carried out a calibration procedure similar to the one proposed in \cite{Brunacci2023_1} to eliminate the bias present in the distance measurements. 
More specifically, we calibrate \textbf{SUL-IS} according to the specific configuration of each experiment, while for \textbf{AUL-IS} the calibration is performed only once with the \textit{TagBot} placed in front of the \textit{AnchorBot}. 
It should be noted that \textbf{SUL-IS} is favored in the comparison since it is calibrated \textit{ad-hoc} for each configuration.

The trained DRL policy is directly deployed on the real \textit{AnchorBot} without any fine-tuning, and the results against the static baseline are reported in Table \ref{tab:real}. 
The learned controller demonstrates strong generalization capabilities as it achieves remarkable localization performance when real robotic platforms and UWB sensors are involved. 
The numerical experiments show that \textbf{AUL-IS} outperforms \textbf{SUL-IS} in each considered scenario, confirming what we observe in the simulations experiments (see Fig. \ref{fig:realEnvs} for qualitative results). 

\begin{table}[!t]
\small
\centering
\caption{Results of the real-world experiments. The mean $\mu$ the standard deviation $\sigma$ of the  \coloredText{MLE} in cm are reported.}
\label{tab:real}
\resizebox{\columnwidth}{!}{%
\begin{tabular}{c|c|c|c|c|c|c|c|c|c|c|c|c}
\cline{2-13}
\multirow{3}{*}{Method} & \multicolumn{6}{c|}{Static TagBot } & \multicolumn{6}{c}{Dynamic TagBot}\Tstrut\\
\cline{2-13}
& \multicolumn{2}{c|}{Static 0°} & \multicolumn{2}{c|}{Static 90°} & \multicolumn{2}{c|}{Static 180°} & \multicolumn{2}{c|}{Straight Line} & \multicolumn{2}{c|}{Circle} & \multicolumn{2}{c}{Square} \Tstrut\\
\cline{2-13}
& $\mu$ & $\sigma$ &$\mu$ & $\sigma$ & $\mu$ & $\sigma$ &$\mu$ & $\sigma$ & $\mu$ & $\sigma$ &$\mu$ & $\sigma$ \\
\hline
SUL-IS
& 5.6 & 3.7 
& 9.0 & 6.3 
& 7.6 & 9.7  
& 32.5 & 22.2  
& 13.2 & 8.8  
& 26.6 & 22.9 \Tstrut\\ 
AUL-IS (Our)
& \textbf{4.9} & \textbf{3.3}  
& \textbf{7.1} & \textbf{4.9}  
& \textbf{5.9} & \textbf{4.2}  
& \textbf{12.7} & \textbf{8.1}  
& \textbf{12.2} & \textbf{7.2}  
& \textbf{14.7} & \textbf{7.8} \\ 
\hline
\end{tabular}%
}
\vspace{-1mm}
\vspace{-3ex}
\end{table}

\section{Conclusion and Future Works} \label{sec:conclusion}
In this work, we presented an infrastructure-less active relative localization method for multi-robot systems using UWB distance sensors. 
Specifically, we proposed a new anchor configuration through a GDOP analysis and introduced a novel loss function used to train a DRL-based controller to adapt the position of the \textit{AnchorBot} and minimize the relative localization error of the \textit{TagBot}.
Extensive simulations and real-world experiments validated the effectiveness of our approach and highlighted its potential for practical deployment.
In particular, the proposed strategy exhibits up to a 60\% error reduction in accuracy compared to existing approaches. 
In future work, we will enhance our method by incorporating sensor fusion techniques to integrate our UWB-based localization approach with visual information. In addition, more challenging robotic platforms, such as Micro Aerial Vehicles (MAVs) will be considered.

\balance

\bibliographystyle{IEEEtran}
\bibliography{_References/Active_Perception}

\end{document}